\documentclass[10pt,twocolumn,letterpaper]{article}

\usepackage{cvpr}
\usepackage{times}
\usepackage{epsfig}
\usepackage{graphicx}
\usepackage{amsmath}
\usepackage{amssymb}

\usepackage{subfiles}
\usepackage{color}
\usepackage{array,multirow}
\usepackage{rotating}
\usepackage{pdflscape}

\newcommand{\vtext}[1]{\rotatebox[origin=b]{90}{#1}}

\cvprfinalcopy 

\usepackage{etoolbox}
\usepackage{bm}
\usepackage{amsthm}
\usepackage{algorithm}
\usepackage{algorithmic}
\usepackage{subfigure}

\newcommand{\MyMapTemplatePrefixc}[4]{\expandafter#1\csname#3#4\endcsname{#2{#4}}} 
\forcsvlist{\MyMapTemplatePrefixc {\def} {\mathcal}{c}} {A,B,C,D,E,F,G,H,I,J,K,L,M,N,O,P,Q,R,S,T,U,V,W,X,Y,Z}  

\newcommand{\MyMapTemplatePrefixtb}[5]{\expandafter#1\csname#4#5\endcsname{#2{#3{#5}}}} 
\forcsvlist{\MyMapTemplatePrefixtb {\def} {\tilde}{\mathbf}{t}} {A,B,C,D,E,F,G,H,I,J,K,L,M,N,O,P,Q,R,S,T,U,V,W,X,Y,Z}  
\forcsvlist{\MyMapTemplatePrefixtb {\def} {\tilde}{\mathbf}{t}} {0,1,a,b,c,d,e,f,g,h,i,j,k,l,m,n,o,p,q,r,s,u,v,w,x,y,z}  

\newcommand{\MyMapTemplateNoPrefix}[3]{\expandafter#1\csname#3\endcsname{#2{#3}}}
\forcsvlist{\MyMapTemplateNoPrefix {\def} {\mathbf} } {0,1,a,b,c,d,e, f, g, h, i, j, k, l, m, n, o, p, q, r, u, v, w, x, y, z} 
\forcsvlist{\MyMapTemplateNoPrefix {\def} {\mathbf} } {A,B,C,D,E,F,G,H,I,J,K,L,M,N,O,P,Q,R,S,T,U,V,W,X,Y,Z}  

\def\etal{\emph{et al.}\@\xspace}

\usepackage{booktabs} 
\usepackage[table,dvipsnames]{xcolor}
\definecolor{rowblue}{RGB}{220,230,240}

\def\cvprPaperID{0000} 

\ifcvprfinal\fi
\begin{document}

\title{ROAD: Reality Oriented Adaptation for Semantic Segmentation of Urban Scenes}

\author{Yuhua Chen$^1$\hspace{10mm}Wen Li$^1$\hspace{10mm}Luc Van Gool$^{1,2}$\\[2mm]
$^1$Computer Vision Laboratory, ETH Zurich\hspace{10mm}
$^2$VISICS, ESAT/PSI, KU Leuven\\[-1.5pt]
{\tt\small \{yuhua.chen,liwen,vangool\}@vision.ee.ethz.ch}
}

\maketitle

\begin{abstract}

Exploiting synthetic data to learn deep models has attracted increasing attention in recent years. However, the intrinsic domain difference between synthetic and real images usually causes a significant performance drop when applying the learned model to real world scenarios. This is mainly due to two reasons: 1) the model overfits to synthetic images, making the convolutional filters incompetent to extract informative representation for real images; 2) there is a distribution difference between synthetic and real data, which is also known as the domain adaptation problem. To this end, we propose a new reality oriented adaptation approach for urban scene semantic segmentation by learning from synthetic data. First, we propose a target guided distillation approach to learn the real image style, which is achieved by training the segmentation model to imitate a pretrained real style model using real images. Second, we further take advantage of the intrinsic spatial structure presented in urban scene images, and propose a spatial-aware adaptation scheme to effectively align the distribution of two domains. These two modules can be readily integrated with existing state-of-the-art semantic segmentation networks to improve their generalizability when adapting from synthetic to real urban scenes. We evaluate the proposed method on Cityscapes dataset by adapting from GTAV and SYNTHIA datasets, where the results demonstrate the effectiveness of our method.
\end{abstract}

\vspace{-2mm}
\section{Introduction}
With the exciting vision of autonomous driving, semantic segmentation of urban scenes, as a key module, has gained increasing attention from both academia and industry. However, collecting and labeling training data for semantic segmentation task is a laborious and expensive process, as it requires per-pixel annotation. This issue becomes even more severe with the surge of deep learning techniques, which usually require a large amount of training data. Therefore, it becomes much desired to exploit low cost ways to acquire data for semantic segmentation.

One way that becomes recently prevalent is to collect photo-realistic synthetic data from video games, where pixel-level annotation can be automatically generated at a much lower cost. For example, Richter \textit{etal.}\cite{richter2016playing} constructed a large scale synthetic urban scene dataset for semantic segmentation from the GTAV game. While the cost of acquiring training data and annotation is largely reduced, synthetic data still suffers from a considerable domain difference from the real data, which usually leads to a significant performance drop when applying the segmentation model to real world urban scenes~\cite{hoffman2016fcns,zhang2017curriculum}. 

The main reasons are two-fold. First, from the perspective of representation, since the model is trained on synthetic images, the convolutional filters tend to overfit to synthetic style images, making them incompetent to extract informative features for real images. Second, from the distribution perspective, synthetic and real data suffers a considerable distribution mismatch, which makes the model biased to synthetic domain.  

\begin{figure*}
\label{fig:roadnet}
\centering
\includegraphics[width=0.95\textwidth]{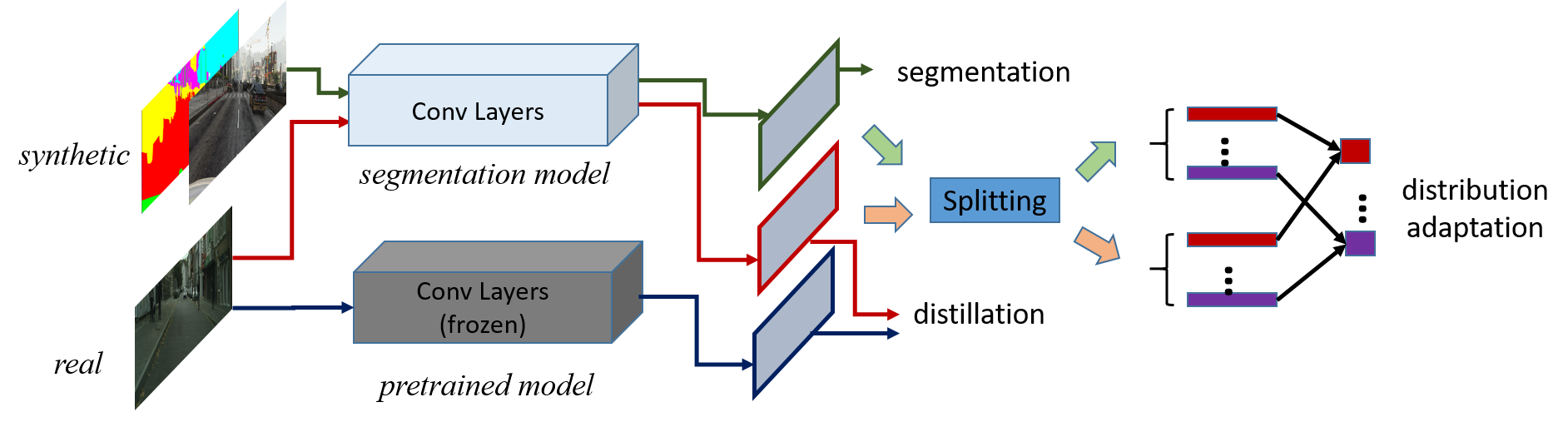}
\caption{Illustration of our Reality Oriented Adaptation networks(ROAD-Net) for semantic segmentation of urban scenes. Our network is built upon conventional semantic segmentation networks, and incorporates a \textit{target guided distillation} module for real style orientation, and a \textit{spatial-aware adaptation} module for real distribution orientation.}
\end{figure*}
    
To overcome such problems, we propose \textbf{R}eality \textbf{O}riented \textbf{AD}aptation Networks(ROAD-Net) for semantic segmentation of urban scenes by learning from synthetic data. We address the above two issues respectively by a target guided distillation module for real style orientation, and a spatial-aware adaptation module for real distribution orientation, which are described respectively as follows,  
\begin{itemize}
\item \textbf{real style orientation: } To prevent the segmentation model from overfitting to  synthetic images, we propose to use the target real images to imitate a pretrained real style model. This can be achieved using the model distillation strategy by enforcing the output from segmentation model similar with the output of a pretrained model. On one hand, this encourages convolutional filters to fit to the real images through the distillation task. On the other hand, it also enforces the segmentation network to preserve good discriminative for real images by approaching the semantic output from the pretrained model. We refer to it as \textit{target guided distillation.} 
\item \textbf{real distribution orientation: } To deal with the distribution mismatch for semantic segmentation, several recent works \cite{hoffman2016fcns,zhang2017curriculum} applied domain adaptation methods on pixel-level features. However, this is generally a challenging task partially due to the large visual variance in urban scene. For example, the objects in the central region are usually much smaller, compared to objects in the outer region. Aligning distributions directly under such large variance presents significant difficulty. We therefore aim to ease the domain adaptation task by exploiting the intrinsic geometry information presented in urban scene. We propose a \textit{spatial-aware adaptation} method to effectively align the two domains while considering the difference in spatial distribution. In particular, we divide the scene image into different spatial regions, and align the source and target domain samples from the same spatial region respectively. In this way, we align features from two domains with similar spatial properties.
\end{itemize}
The two modules above can be easily integrated with existing state-of-the-art semantic segmentation networks to boost their generalizability when adapting from synthetic to real urban scenes. We conduct extensive experiments by using the GTAV dataset and the Cityscapes datset. Our proposed method achieves a new state-of-the-art of $39.4\%$ mean IoU. To further validate the effectiveness, we additionally evaluate our methods using SYNTHIA dataset and the Cityscapes dataset, in which the proposed method outperforms other competing methods as well.

\section{Related Works}

\textbf{Semantic Segmentation:} Semantic segmentation is a highly active field since decades ago with large amount of methods proposed, we briefly review some of the works with a focus on CNN-based methods. Traditional works in semantic segmentation~\cite{shotton2008semantic,tighe2010superparsing,zhang2010semantic} are typically based on manually designed image features. With the recent surge of deep learning~\cite{krizhevsky2012imagenet}, learned representation demonstrated its power in many computer vision tasks. Pioneered by \cite{long2015fully}, the power of CNN has been transferred to semantic segmentation and we have witnessed a rapid boost in semantic segmentation performance. Long \textit{etal.}\cite{long2015fully} formulates semantic segmentation as a per-pixel classification problem. Following the line of FCN-based semantic segmentation. Dilated convolution is proposed by~\cite{yu2015multi} to enlarge the receptive field. DeepLab~\cite{chen2016deeplab} incorporates Conditional random filed(CRF) with CNN to reason about spatial relationship. Recently, Zhao \textit{etal.}\cite{zhao2016pyramid} proposed to use \textit{Pyramid Pooling Module} to encode the global and local context, which achieved state-of-the-arts results on multiple datasets. 

\textbf{Domain Adaptation:} A basic assumption in conventional machine learning is that the training and test data are sampled independently from an identical distribution, or \textit{i.i.d} assumption in short. However, this does not always hold in real world scenarios, which often leads to a significant performance drop on the test data when applying the trained model. Domain adaptation aims to alleviate the impact of such distribution mismatch such that the generalization ability of the learned model can be improved on the target domain. In computer vision, domain adaptation has been widely studied as an image classification problem in computer vision~\cite{kulis2011you,gopalan2011domain,gong2012geodesic,fernando2013unsupervised,sun2015return,long2015learning,ganin2015unsupervised,ghifary2016deep,sener2016learning,panareda2017open,motiian2017unified,li2017domain}. Conventional methods include asymmetric metric learning~\cite{kulis2011you}, subspace interpolation~\cite{gopalan2011domain}, geodesic flow kernel~\cite{gong2012geodesic}, subspace alignment~\cite{fernando2013unsupervised}, covariance matrix alignment~\cite{sun2015return}, \etc. Recent works aim to improve the domain adaptability of deep neural networks, including \cite{long2015learning,ganin2015unsupervised,ghifary2016deep,sener2016learning,panareda2017open,motiian2017unified,li2017deeper,haeusser2017associative,lu2017unsupervised,maria2017autodial}. Different from those works, our work aims to solve the semantic segmentation task, which is more challenging due to the large variance in pixel-level features. 

\textbf{Domain Adaptation for Semantic Segmentation:} So far most works in domain adaptation focus on the task of image classification. Not until recently has the community pay attention to domain shift problem in semantic segmentation. This line of research is pioneered by~\cite{hoffman2016fcns}, where they use adversarial training to align the feature from both domain. Another approach is done by~\cite{zhang2017curriculum}, where they use a curriculum learning style approach to solve the domain shift. There are also some concurrent works~\cite{sankaranarayanan2017unsupervised,saito2017maximum,chen2017no,tsai2018learning} concerning similar problem in different ways.

\textbf{Learning Using Synthetic Data: } There are also a few works proposed to learn from synthetic data \cite{sun2014virtual,xu2016hierarchical,ros2016synthia,vazquez2014virtual,peng2017synthetic,shrivastava2016learning}. In \cite{sun2014virtual}, generic object detectors are trained from synthetic images, while in \cite{vazquez2014virtual} virtual images are used to improve pedestrian detections in real environment. More recently, \cite{chen2018domain} extends Faster R-CNN to learn a domain-invariant detector using synthetic data or data from another domain.

\textbf{Other Related Works :} From the methodology, our work is inspired by the model distillation~\cite{hinton2015distilling}, which was proposed for network compression. The following up works employed a similar strategy to perform knowledge transfer \cite{li2016learning,Wang2017}, and cross-modality supervision transfer \cite{gupta2016cross}. Our work, on the other hand, using the distillation strategy for learning the real style convolutional filters for semantic segmentation. Our work is also inspired by the previous works on scene understanding \cite{lazebnik2006beyond,hoiem2008putting}, in which they also exploited the intrinsic spatial layout prior of scene images for different tasks in various ways. We share a similar philosophy with those work, but aim to solve the distribution alignment between two domains for the urban scene semantic segmentation, which results in a totally different methodology. 

\section{Reality Oriented Adaptation Networks}
In this section, we present our Reality Oriented Adaptation networks(ROAD-Net) for urban scene semantic segmentation. Two new modules are proposed in our ROAD-Net model, the \textit{target guided distillation} and \textit{spatial-aware adaptation}, which will be introduced respectively in this section.

\subsection{Target Guided Distillation}
\label{sec:rso}
When learning a model from synthetic urban scenes, one major issue is that synthetic data usually exhibits a clear visual difference with real images. As a result, the learned model is biased towards the synthetic data, which leads to an unsatisfactory segmentation performance on real urban scene images. 

From the perspective of representation learning, this is largely because that the convolutional filters in the semantic segmentation models overfit to the synthetic image style. Consequently, when taking a real image as input, the feature representation generated using the learned model might not be sufficiently informative for semantic segmentation, since the convolutional filters tend to extract information that look discriminative in synthetic style only. To cope with this issue, we therefore propose to use the target real images to guide the segmentation model to learn robust convolutional filters for the real urban scenes. 

\begin{figure}
\centering
\includegraphics[width=0.9\linewidth]{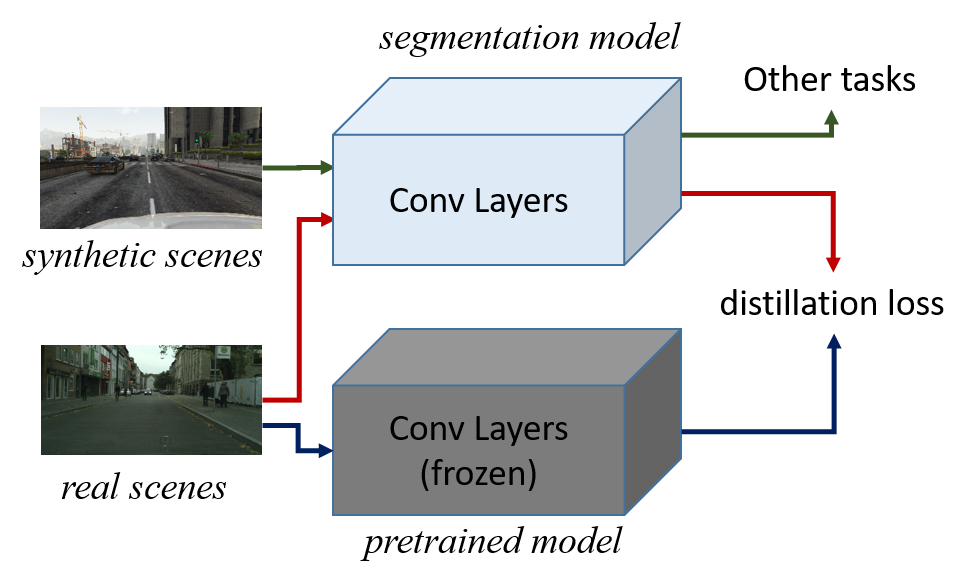}
\caption{Illustration of target guided distillation. By feeding the target real images into both the segmentation model, and a pretrained model, the segmentation model is encourage to imitate the pretrained model to learn real style convolutional filters.}
\label{fig:fig_rso}
\end{figure}

Our method is motivated by the common practice in computer vision community of initializing network weights using a network~\cite{he2016deep,simonyan2014very} pretrained on large-scale image dataset such as ImageNet~\cite{imagenet_cvpr09}. Such strategy has also been widely exploited for learning semantic segmentation models \cite{chen2016deeplab,yu2015multi,zhao2016pyramid}. Recall that the original ImageNet model is pretrained on real images, so we propose to employ a distillation loss to guide the semantic segmentation models to behave like the pretrained real style model. 

We illustrate the target guided distillation process in Figure~\ref{fig:fig_rso}. The source synthetic urban scene images and corresponding annotation are used to learn the semantic segmentation model. At the same time, the unlabeled real urban scene images from target domain are used to imitate their corresponding feature map output from the pretrained ImageNet model. The pretrained model has the same structure as the backbone network used in the semantic segmentation model, and is frozen during this process. 

Formally, for a real image, let us denote $\x_{i,j}$ as the activation at the position $(i,j)$ of the feature map from the semantic segmentation model, and also denote $\z_{i,j}$ as an activation at the same location of the feature map from the pretrained model, then the loss for target guided distillation can be written as,
\begin{eqnarray}
\label{eqn:loss_rso}
\cL_{dist} = \frac{1}{N}\sum_{i,j}\|\x_{i,j} - \z_{i,j}\|_2
\end{eqnarray}
where $N$ is the number of activations at the feature map, and $\|\cdot\|_2$ is the Euclidean distance. 

\textbf{Discussion:} There also exists other alternative approaches to prevent convolutional filters from overfitting to synthetic data. For instance, considering the semantic segmentation model is initialized with a pretrained ImageNet model which is learned from the real images, a possible way is to freeze a few convolutional layers (\ie, to set the learning rates for those layers to zero), such that convolutional filters will not be corrupted by synthetic data. As shown in our experiments(\textit{c.f.} Section~\ref{sec:exp_rso}), this does work to some extent. However, it is still inferior to our target guided distillation approach. The main reason is that the ImageNet model is trained for image classification, and may not be optimal to the semantic segmentation of urban scenes. Freezing a few layers may help to prevent the model from overfitting to synthetic data, but also limits its capacity for the semantic segmentation task. As a comparison, in target guided distillation, all weights are allowed to be tuned, and convolutional filters are guided in a soft manner to imitate the ImageNet model when being trained for the segmentation task. 

Another alternative approach is to use source data to perform the distillation task, which is also known as learning without forgetting~\cite{li2016learning}. Though it is able to imitate the ImageNet pretrained model, we argue that it is not as effective as  using the target real urban scene images, because the convolutional filters can be still corrupted due to taking solely synthetic images as input. Moreover, the deep neural networks often contain multiple layers, and the filters in higher layers are able to be trained to well imitate the ImageNet model even with corrupted low layers convolutional filters. We conduct an experimental comparison with all alternative baselines in Section~\ref{sec:exp_rso}. 

\subsection{Spatial-Aware Adaptation}
Even with the target guided distillation, the feature presentation of two domains extracted by the semantic segmentation networks could still suffer from a distribution mismatch, due to the large domain difference between synthetic and real images. Thus, it is desirable to align the distributions of two domains with domain adaptation methods. 

However, aligning pixel-level features between synthetic and real data for urban scene images is non-trivial. Conventional domain adaptation approaches are usually proposed for the image classification task. While similar methodology can be applied by taking each pixel-level feature as a training sample, it is still challenging to fully reduce the distribution mismatch, since the pixels vary significantly in appearance~\cite{hoffman2016fcns} and scale~\cite{chen2016scale}.

To address this issue, we hence propose to leverage the intrinsic spatial structure in urban scene images to improve the domain distribution alignment between synthetic and real data. Our motivations are two-fold. Firstly, while the pixel-level features vary a lot, such variance also exhibits a nice spatial pattern. For example, in an urban scene image, the objects in the central region are usually small, while objects in the outer region are relatively larger. Second, the semantic categories also roughly follow a spatial layout. Usually \textit{road} appears in the bottom part of an image, and \textit{sky} appears on the top part of an image. Therefore, it is beneficial to align pixel-level features with similar sizes and semantics based on their spatial location.  

\begin{figure}
\centering
\includegraphics[width=1.0\linewidth]{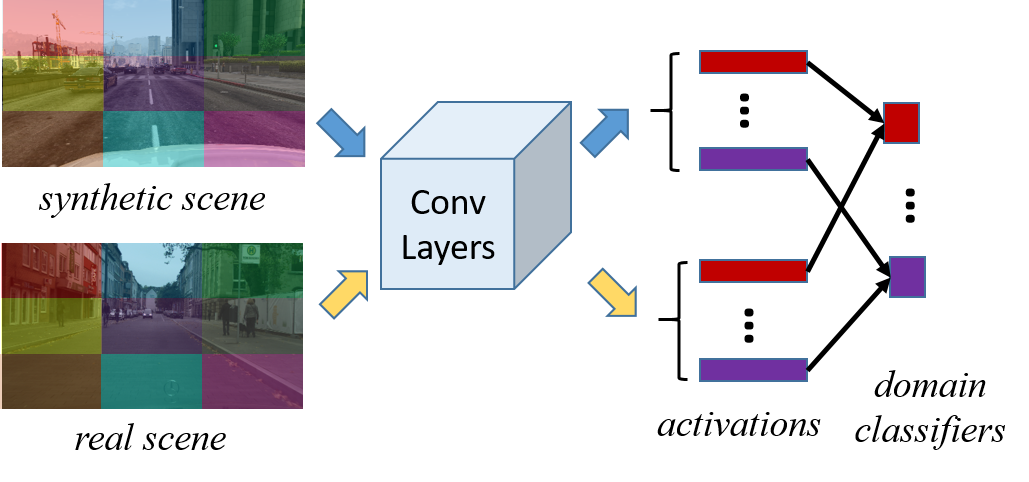}
\caption{Illustration of spatial-aware adaptation. The source and target pixel-level features in the same spatial region are aligned(\ie, the regions marked with the same color), which eases the domain distribution adaptation task. }
\label{fig:fig_rdo}
\vspace{-4mm}
\end{figure}

In particular, we divide each scene image into different spatial regions, and domain alignment is performed on the pixel-level features from the same spatial region. We illustrate this process in Figure~\ref{fig:fig_rdo}. For each region, any traditional domain distribution alignment can be deployed. Formally, suppose there are $m = 1, \ldots, M$ regions. We denote $\cR_m$ as the set of indices contains in the $m$-th region, \ie, $(u,v) \in \cR_m$ means that a pixel-level activation $\x_{u,v}$ locates at this region, and vice versa. Let us denote $\cX^s_m = \{\x^s_{u,v} | (u,v) \in \cR_m\}$ and $\cX^t_m = \{\x^t_{u,v} | (u,v) \in \cR_m\}$ as all pixel-level activations locate at the $m$-th region for source and target domains respectively, then the loss for spatial-aware adaptation can be written,
\begin{eqnarray}
\label{eqn:loss_rdo}
\cL_{spt} = \sum_{m=1}^M \cL_{da}(\cX^s_m, \cX^t_m)
\end{eqnarray}
where $\cL_{da}$ is a domain adaptation loss which measures the domain difference between two sets of samples. We present an example of $\cL_{da}$ below.

\textbf{Domain Adversarial Training: } Aligning two distributions has been widely studied in the literature. In this work, we deploy an $\cH$-divergence based loss used in the DANN model~\cite{ganin2015unsupervised}. Specifically, let us denote $h : \x\rightarrow\{0,1\}$ as a domain classifier, which is used to predict which domain an input pixel-level feature $\x$ comes from, where $0$ denotes the source domain, and $1$ denotes the target domain. Given a set of training data $\cX = \cX^s \cup \cX^t$, the loss for training the domain classifier $h$ can be written as:
\begin{eqnarray}
\label{eqn:loss_dc}
\cL_{\cH}(\cX^s, \cX^t) = \frac{1}{|\cX|}\sum_{\x \in \cX}\ell(h(\x), d)
\end{eqnarray}
where $|\cX|$ is the number of samples in $\cX$, $d \in \{0, 1\}$ is the domain label of $\x$, $\ell(\cdot,\cdot)$ is a conventional classification loss for which we use the cross-entropy loss. 

Intuitively, training a domain classifier is to distinguish samples from two domains. To reduce the domain difference, we thus encourage the activation $\x$ to be domain-indistinguishable. Considering each $\x$ is generated from a base network, denoted by $F$, we thus need to optimize $F$ such that the domain classification loss $\cL_{\cH}$ is maximized. By jointly learning the domain classifier $h$ and the base network $F$, we arrive at the following maxmin problem, 
\begin{eqnarray}
\label{eqn:loss_dann}
\max_{F}\min_{h} \cL_{\cH}(\cX^s, \cX^t)
\end{eqnarray}
The above maxmin problem can be optimized in an adversarial training manner. Hoffman \etal~\cite{hoffman2016fcns} implemented it by switching domain label similarly as in the Generative Adversarial Networks (GAN)~\cite{goodfellow2014generative}. We follow \cite{ganin2015unsupervised} to insert a gradient reverse layer between $F$ and $h$. Particularly, in back-propagation, the sign of gradients will be flipped when being passed through the gradient reverse layer, hence the classifier $h$ is minimized while the network $F$  maximized by directly using the conventional optimization methods like stochastic gradient descent (SGD). 

\subsection{Network Overview}

The aforementioned two modules can be integrated with a conventional semantic segmentation network, such as DilatedNet~\cite{yu2015multi}, DeepLab~\cite{chen2016deeplab}, PSPNet~\cite{zhao2016pyramid}~\etc. In particular, let us denote $\cL_{seg}$ as the segmentation loss, then our ROAD-Net model can be trained by minimizing a joint loss as follows, 
\begin{eqnarray}
\cL_{ROAD} = \cL_{seg} + \lambda_1 \cL_{dist} + \lambda_2\cL_{spt}. 
\end{eqnarray}
where $\lambda_1$ and $\lambda_2$ are two trade-off parameters, which are set to $0.1$ and $0.01$ in our experiments. Note that the loss $\cL_{spt}$ in the above equation corresponds to $\cL_{\cH}$ in (\ref{eqn:loss_dann}). Although we have removed the $\max$ operator in (\ref{eqn:loss_dann}) from the above loss, it can be automatically achieved with the reverse gradient layer. 

We illustrate the pipeline of our proposed ROAD-net in Fig~\ref{fig:roadnet}. During training, both synthetic and real images are fed into the network as input. The synthetic images and the annotation are used to train the segmentation task, while the real images are used to train the target guided distillation task. Both synthetic and real images are used for optimizing the spatial-aware adaptation loss. Note that, when training semantic segmentation model, due to the high resolution of input image, the input image is usually randomly cropped to fit the GPU memory. To perform the spatial-aware adaptation, we therefore build a spatial-aware splitting layer, in which we recover the location for each activation in the original image coordinate, and split them into different domain classifiers according to the region it comes from. During the test phase, the two newly added modules can be removed, and one can perform semantic segmentation as the same as for the conventional semantic segmentation models.

\begin{table*}[h]
\center
\resizebox{\textwidth}{!}{%
\setlength{\tabcolsep}{3pt}
\begin{tabular}{ c| c |c c| c c c c c c c c c c c c c c c c c c  c | c}
\hline
& & dst & spt & \vtext{road}& \vtext{sidewalk} & \vtext{building} & \vtext{wall} & \vtext{fence} & \vtext{pole} & \vtext{traffic light} & \vtext{traffic sign} & \vtext{vegetation} & \vtext{terrain} & \vtext{sky} & \vtext{person} & \vtext{rider} & \vtext{car} & \vtext{truck} & \vtext{bus} & \vtext{train} & \vtext{motorbike} & \vtext{bicycle} & \vtext{mean IoU} \\ \hline
\multirow{4}{*}{DeepLab}& NonAdapt & & & 29.8 & 16.0 & 56.6 &  9.2 &  17.3 &  13.5 & 13.6 & 9.8 & 74.9 & 6.7 & 54.3 & 41.9 &  2.9 &  45.0 & 3.3 &  13.1 & 1.3 & 6.8 & 0.0 & 21.9 \\ \cline{2-24}
& \multirow{3}{*}{Ours} & \checkmark& &  82.0  &29.5  &73.8 &20.0  &12.7  &8.5  &14.6  &9.0  &73.2  &27.6  &63.2  &39.3  &5.9  &71.7  &23.1  &17.6  &\textbf{14.4}  &4.0 &0.0  & 31.1  \\  \cline{3-24}
 & & & \checkmark& 85.3 & \textbf{36.1}  & 77.5  & 24.1  & 18.8  & 21.4  &22.5  &\textbf{13.2}  &71.3  &20.9  &54.5  &40.6  &6.9  &76.7  &\textbf{19.3}  &16.4  &7.0  &7.1  &0.0  &32.6  \\  \cline{3-24}
& & \checkmark & \checkmark & \textbf{85.4}  & 31.2  & \textbf{78.6}  & \textbf{27.9}  & \textbf{22.2}  & \textbf{21.9}  & \textbf{23.7}  & 11.4  & \textbf{80.7}  & \textbf{29.3} & \textbf{68.9} & \textbf{48.5} & \textbf{14.1}  & \textbf{78.0}  & 19.1  & \textbf{23.8}  & 9.4  & \textbf{8.3}  & 0.0 & \textbf{35.9}  \\
\hline
\hline
\multirow{4}{*}{PSPNet} &NonAdapt &  &  & 14.6  & 17.0  & 40.4  & 21.1  & 14.7  & 18.3 & 18.0 & 5.6 & 80.8 & 16.0 & 68.4 & 49.1 & 4.4 & 78.5 &\textbf{31.8}  & 23.9 & 1.5 & 24.4 & 0.0 & 27.8  \\ \cline{2-24}
& \multirow{3}{*}{Ours} & \checkmark& & 73.2  & 32.0  & 71.0  & 25.2  & \textbf{31.2}  & 15.7  & 14.4  & \textbf{15.2}   & \textbf{82.5} & \textbf{36.1} & 73.1 & 52.1 & 17.1 & 78.1  & 26.5 & 28.0 & 3.1 & 24.7 & 4.4 &  36.8  \\  \cline{3-24}
 & & & \checkmark& 75.9 & 34.2 & \textbf{72.3} & 28.0 & 21.7 & \textbf{30.4} & 25.2 & \textbf{15.2} & 79.4 & 33.5 &\textbf{73.5} & 52.3 & \textbf{18.7} & 76.2 & 31.0 & 25.2 & 3.5 & 23.7 & 2.1  & 37.8 \\  \cline{3-24}
& & \checkmark & \checkmark & \textbf{76.3}  & \textbf{36.1} &  69.6 & \textbf{28.6} & 22.4 & 28.6 & \textbf{29.3} & 14.8 & 82.3  & 35.3  & 72.9 & \textbf{54.4} & 17.8 & \textbf{78.9} & 27.7 & \textbf{30.3} & \textbf{4.0} & \textbf{24.9} & \textbf{12.6} & \textbf{39.4} \\
\hline
\end{tabular}
}
\vspace{2mm}
\caption{The segmentation results on the Cityscapes dataset by using the GTAV dataset as the source domain. DeepLab and PSPNet are used as the base model, respectively. We report different variants of our proposed ROAD-Net, where ``dst'' and ``spt'' in the head row refers to the \textit{target guided distillation} module and the \textit{spatial-aware adaptation} module, respectively. The ``NonAdapt'' refers to the vanilla model which is trained using the GTAV data only. The best results for each base model are denoted in bold. }
\vspace{-3mm}
\label{tab:main_results}
\end{table*}

\section{Experimental Results}
\label{sec:exp}
In this section, we present our experimental results on semantic segmentation of real urban scenes by learning from synthetic data. Experimental analysis and comparison with state-of-the-arts are also provided. 

\subsection{Experiment Setup}
We follow the classical unsupervised domain adaptation protocol where the supervision is assumed to be given in the source domain, and only unlabeled data is provided in the target domain. Following previous works~\cite{hoffman2016fcns,zhang2017curriculum}, the experimental validation are conducted on the GTAV dataset~\cite{richter2016playing} and Cityscapes dataset~\cite{Cordts2016Cityscapes}. We use GTAV dataset as our source domain, and we have access to the pixel-level annotation, and we use Cityscapes dataset as the target domain. Our goal is to learn a semantic segmentation model from synthetic data. We briefly introduce the datasets used in our experiment in below:

\vspace{-4mm}
\paragraph{Cityscapes} is a dataset focused on autonomous driving, which consists of $2,975$ images in training set, and $500$ images for validation. The images have a fixed resolution of $2048\times 1024$ pixels, and are captured by a video camera mounted in front of a car. 19 semantic categories are provided with pixel-level labels. In our experiment, we use the unlabeled images from the training set as the target domain to adapt our segmentation model, and the results are reported on the validation set. 

\vspace{-4mm}

\paragraph{GTAV} is a dataset recently proposed for learning from synthetic data. It has $25,000$ photo-realistic images rendered by the gaming engine Grand Theft Auto V (GTAV). The resolution of images is around $2000\times 1000$ pixels which is similar with Cityscapes, the semantic categories are also compatible between the two datasets.

In our experiments, we test the our proposed ROAD-Net with two semantic segmentation methods: DeepLab v2~\cite{chen2016deeplab} and PSPNet~\cite{zhao2016pyramid}. However, as discussed in the previous section, our method can be straightforwardly applied to other semantic segmentation methods as well. 

The network is initialized using an ImageNet pre-trained weights, and the last convolutional layer is replaced with our new classification head to predict Cityscapes label. Similar with \cite{chen2016deeplab}, a learning rate of $2.5\times 10^{-4}$ is used, and learning rate policy of \textit{poly} is used in our experiment. Each batch contains 10 sampled patches of size $321\times 321$, of which $5$ patches are from source domain, and the other $5$ patches from target domain. The cross-entropy loss is used for supervising semantic segmentation in the source domain as in~\cite{long2015fully,chen2016deeplab}.

\subsection{Experimental Results}
\label{sec:exp_results}
For a comprehensive study on our method, we include two variants of our method, using only the \textit{target guided distillation} module (referred to as ``dst") and using  only the \textit{spatial-aware adaptation} module (referred to as ``spt"). The vanilla base network without using any module is also included as a baseline (referred to as ``NonAdapt"). Two different semantic segmentation networks are used, DeepLab V2 and the PSP Net, where the former one uses VGG-16 as the backbone network, and the latter one uses ResNet-101. 

\begin{figure*}[h]
\small
\centering
      
      \resizebox{\textwidth}{!}{%
	  \setlength{\fboxsep}{0pt}
      \fbox{\includegraphics[height=0.1\textwidth]{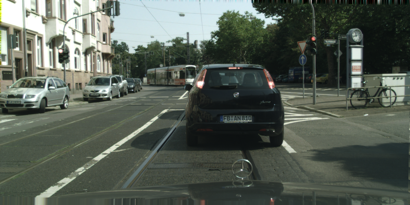}}
      \fbox{\includegraphics[height=0.1\textwidth]{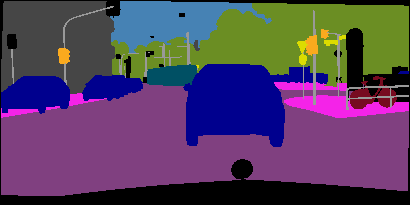}}
      \fbox{\includegraphics[height=0.1\textwidth]{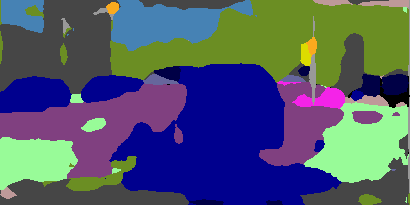}}
      \fbox{\includegraphics[height=0.1\textwidth]{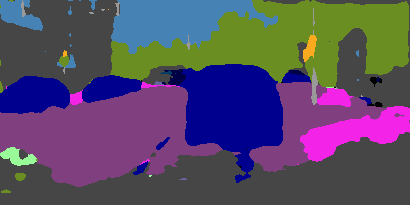}}
      \fbox{\includegraphics[height=0.1\textwidth]{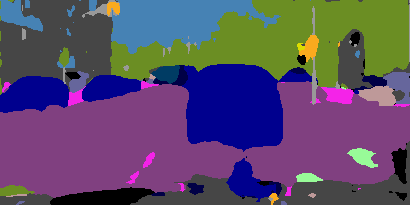}}
      \fbox{\includegraphics[height=0.1\textwidth]{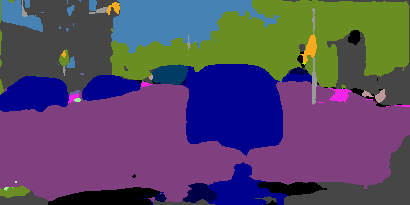}}
      }\\

      \resizebox{\linewidth}{!}{%
	  \setlength{\fboxsep}{0pt}
      \fbox{\includegraphics[height=0.1\textwidth]{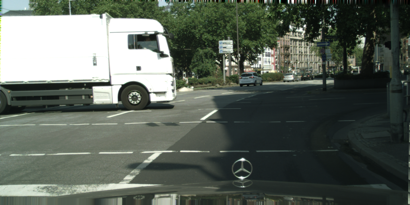}}
      \fbox{\includegraphics[height=0.1\textwidth]{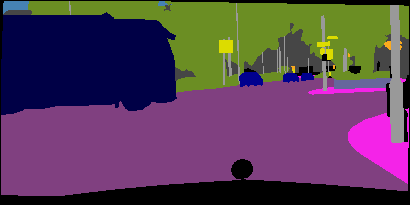}}
      \fbox{\includegraphics[height=0.1\textwidth]{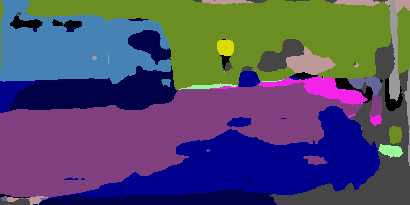}}
      \fbox{\includegraphics[height=0.1\textwidth]{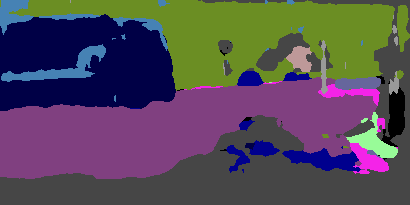}}
      \fbox{\includegraphics[height=0.1\textwidth]{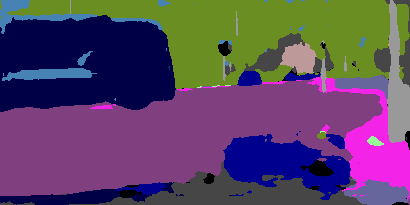}}
      \fbox{\includegraphics[height=0.1\textwidth]{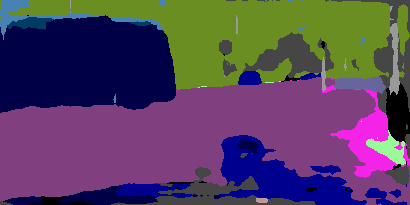}}
      }\\

      \resizebox{\linewidth}{!}{%
	  \setlength{\fboxsep}{0pt}
      \fbox{\includegraphics[height=0.1\textwidth]{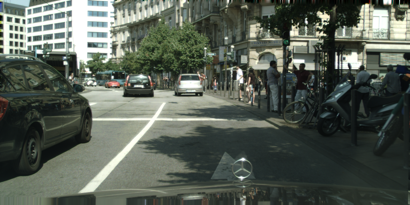}}
      \fbox{\includegraphics[height=0.1\textwidth]{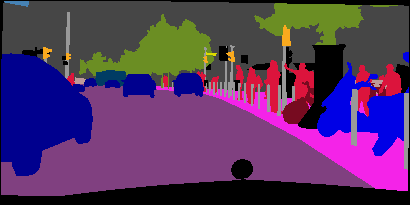}}
      \fbox{\includegraphics[height=0.1\textwidth]{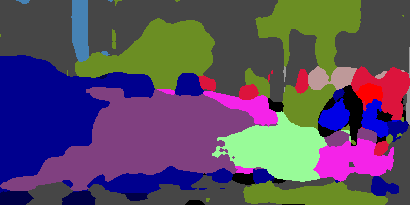}}
      \fbox{\includegraphics[height=0.1\textwidth]{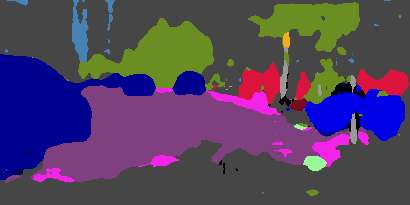}}
      \fbox{\includegraphics[height=0.1\textwidth]{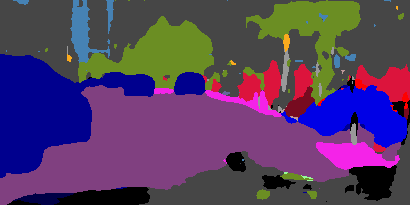}}
      \fbox{\includegraphics[height=0.1\textwidth]{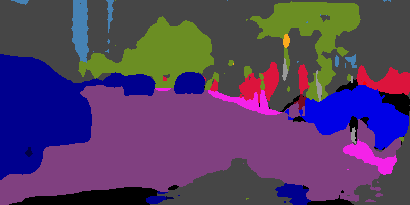}}
      }\\

      \resizebox{\linewidth}{!}{%
	  \setlength{\fboxsep}{0pt}
      \fbox{\includegraphics[height=0.1\textwidth]{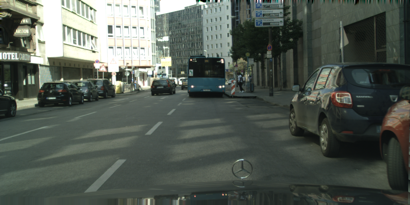}}
      \fbox{\includegraphics[height=0.1\textwidth]{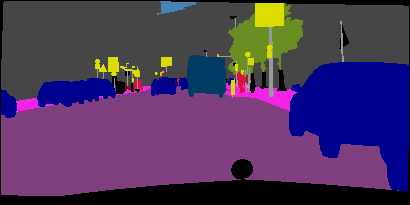}}
      \fbox{\includegraphics[height=0.1\textwidth]{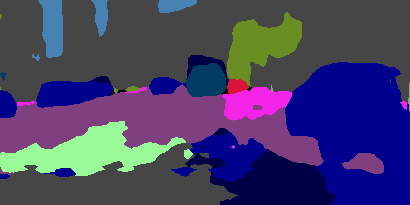}}
      \fbox{\includegraphics[height=0.1\textwidth]{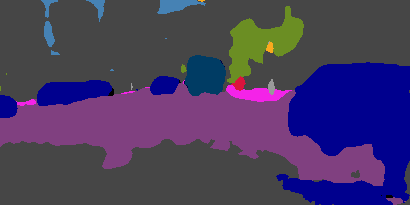}}
      \fbox{\includegraphics[height=0.1\textwidth]{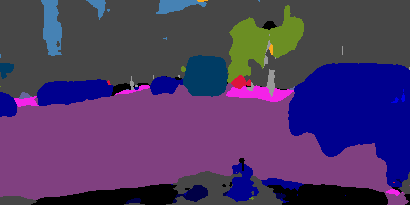}}
      \fbox{\includegraphics[height=0.1\textwidth]{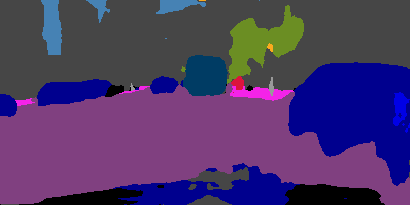}}
      }\\

      \resizebox{\linewidth}{!}{%
	  \setlength{\fboxsep}{0pt}
      \fbox{\includegraphics[height=0.1\textwidth]{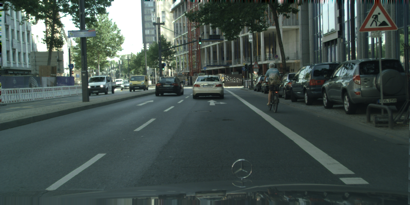}}
      \fbox{\includegraphics[height=0.1\textwidth]{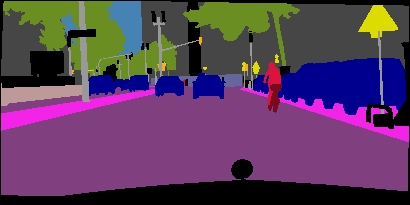}}
      \fbox{\includegraphics[height=0.1\textwidth]{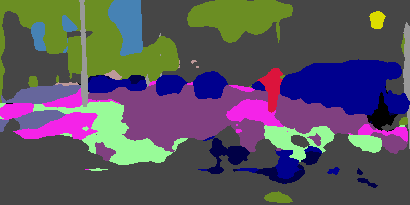}}
      \fbox{\includegraphics[height=0.1\textwidth]{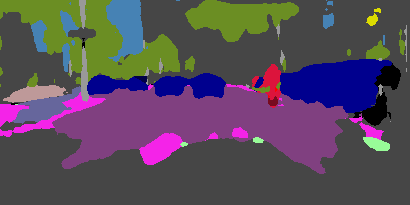}}
      \fbox{\includegraphics[height=0.1\textwidth]{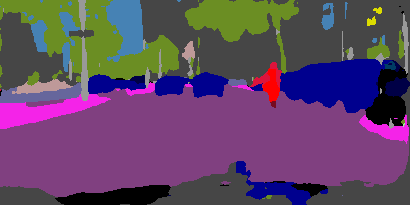}}
      \fbox{\includegraphics[height=0.1\textwidth]{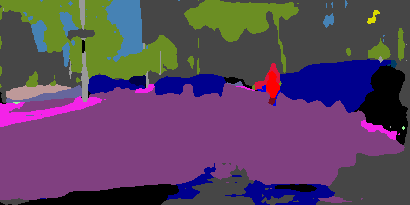}}
      }\\

      \resizebox{\linewidth}{!}{%
	  \setlength{\fboxsep}{0pt}
      \fbox{\includegraphics[height=0.1\textwidth]{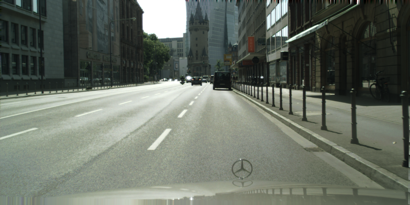}}
      \fbox{\includegraphics[height=0.1\textwidth]{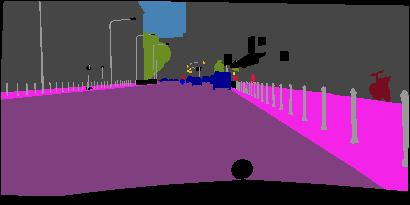}}
      \fbox{\includegraphics[height=0.1\textwidth]{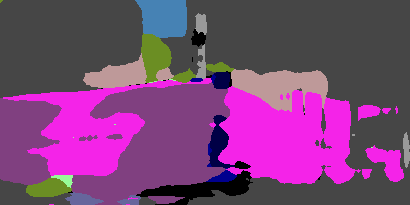}}
      \fbox{\includegraphics[height=0.1\textwidth]{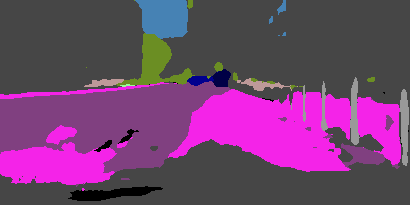}}
      \fbox{\includegraphics[height=0.1\textwidth]{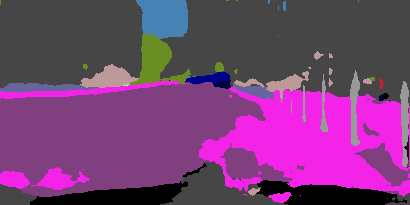}}
      \fbox{\includegraphics[height=0.1\textwidth]{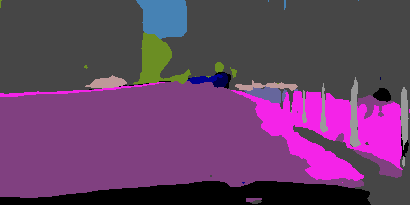}}
      }\\

      \resizebox{\linewidth}{!}{%
	  \setlength{\fboxsep}{0pt}
      \fbox{\includegraphics[height=0.1\textwidth]{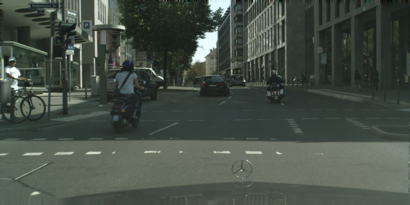}}
      \fbox{\includegraphics[height=0.1\textwidth]{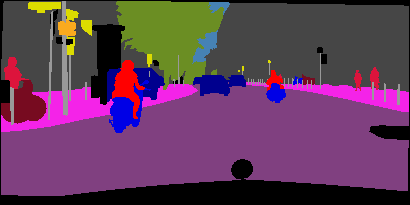}}
      \fbox{\includegraphics[height=0.1\textwidth]{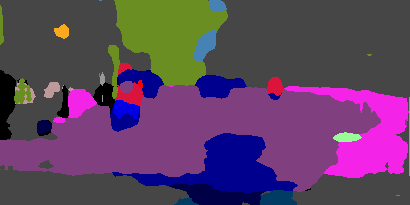}}
      \fbox{\includegraphics[height=0.1\textwidth]{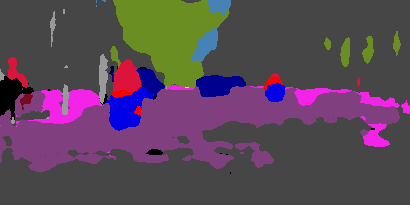}}
      \fbox{\includegraphics[height=0.1\textwidth]{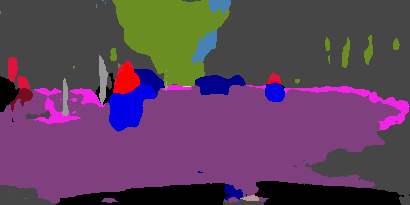}}
      \fbox{\includegraphics[height=0.1\textwidth]{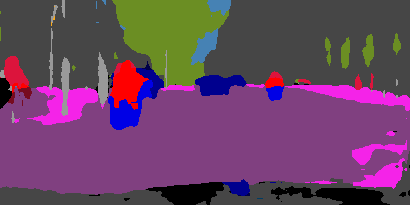}}
      }\\


      \vspace{2mm}
      (a)\hspace{0.15\linewidth}(b)\hspace{0.15\linewidth}(c)\hspace{0.15\linewidth}(d)\hspace{0.15\linewidth}(e)\hspace{0.15\linewidth}(f)
      \vspace{2mm}
        \caption{Qualitative semantic segmentation results on the Cityscapes: (a) original image, (b) ground truth annotation, (c) NonAdapt baseline, (d) our model using target guided distillation only, (e) our model using spatial aware adaptation only, (f) our final ROAD-Net model.}
        \vspace{-2mm}
\label{fig:qualitative}
\end{figure*}

The results of all variants of our methods based on two different models are summarized in Talbe~\ref{tab:main_results}. Taking the results using DeepLab model as an example, the ``NonAdapt" baseline gives $21.9\%$ mean IoU, which is similar to those reported in \cite{hoffman2016fcns,zhang2017curriculum}, where VGG-16 was also used as the backbone network. By using our proposed \textit{target guided distillation} module and \textit{spatial-aware adaptation} module respectively, we obtain a mean IoU of $31.1\%$ and $32.6\%$, which improves the baseline by a large margin of $9.2\%$ and $10.7\%$, respectively. This clearly demonstrates the effectiveness of the two modules for orienting the semantic segmentation model to learn reality.  By combining two modules, our final ROAD-Net gains further improvements, achieving $35.9\%$ in terms of mean IoU. Those observations can also be observed for the results using the PSPNet. Moreover, using PSPNet, all methods are improved, and our ROAD-net finally achieves $39.4\%$ mean IoU. Some randomly selected qualitative examples are shown in Fig. \ref{fig:qualitative}. We observe that the proposed method, especially the spatial aware module, improves the visual segmentation result notably, and produces predictions with much more reasonable spatial layout.

The proposed method does not achieve significant improvement on some categories, such as \textit{pole} and \textit{traffic sign}. A possible reason might be that those categories have very few pixels(\eg in Cityscapes Val, only $1.8M$ pixels for \textit{traffic light}, as a comparison, there are $345M$ pixels for \textit{road}), which makes the results less stable compared to categories with more pixels. 

\begin{figure}
\centering
\includegraphics[width=0.7\linewidth]{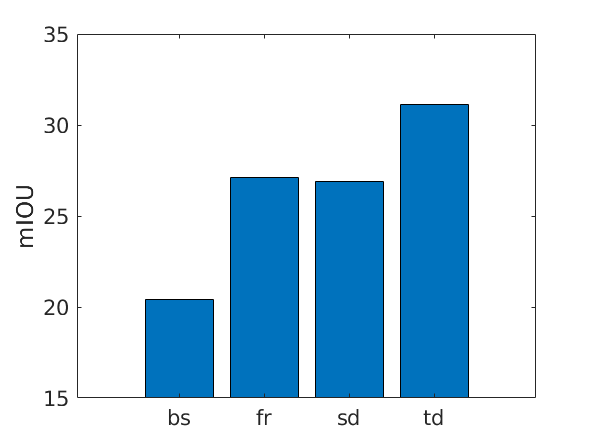}
\caption{Study on alternative methods to target guided distillation: ``bs'' refers to the basline NonAdapt method, ``fr'' refers to the frozen method, ``sd'' refers to the source distillation method, and ``td'' refers to the target guided distillation.}
\label{fig:frozen}
\end{figure}

\subsection{Analysis on Real Style Orientation}
\label{sec:exp_rso}

To validate the effectiveness of our proposed target guided distillation module, we conduct experiments by comparing with the two alternative methods discussed in Section~\ref{sec:rso}. In particular, the ``\textbf{frozen}'' method refers to freezing the first a few layers when training the segmentation model, and ``$\textbf{source distillation}$" refers to replacing real images with synthetic images for distillation. For all methods, the DeepLab network is deployed due to its efficiency. All other experimental settings are identical with previous experiments. 

The results of different methods are summarized in Fig~\ref{fig:frozen} where the ``NonAdapt" baseline is also included for comparison. From the figure, we observe that all methods achieve performance gain when compared with the ``NonAdapt'' baseline, which implies that it is the beneficial to prevent the model from overfitting the synthetic images. Directly freezing a few layers is similar to the ``source distillation' method. Our target guided distillation method achieves the best improvement, which demonstrates its effectiveness by using real images in distillation to prevent the models from over-fitting synthetic style images. 

\begin{figure}
\centering
\includegraphics[width=0.7\linewidth]{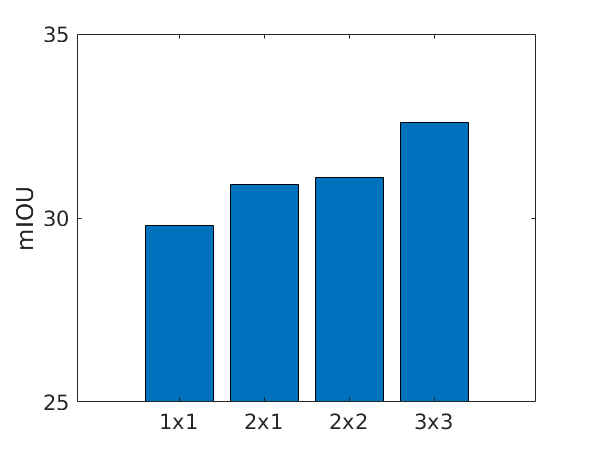}
\caption{Effects of different partitions in spatial-aware adaptation. $H\times W$ denotes that the image is divided $H$ times along the vertical axis, and $W$ times along the horizontal axis.}
\label{fig:spatial}
\vspace{-4mm}
\end{figure}

\subsection{Analysis on Real Distribution Orientation}
\label{sec:exp_rdo}
In our experiments, we divide the image into $3 \times 3$ spatial regions for our spatial-aware adaptation module. To study the effect of different partitions of regions on our method, we conduct additional experiments by using $1 \times 1$, $2\times 1$, $2\times 2$ partitions, respectively. The DeepLab network is used in this experiment, while other experimental settings remain unchanged with previous experiments. 

The results by using different region partitions are shown in Fig.~\ref{fig:spatial}. Among those, the case of $1 \times 1$ is the ordinary domain adaptation method without considering the spatial information, which is also equivalent to the \textit{global alignment} method in~\cite{hoffman2016fcns}. We observe that all other partition methods outperform this baseline, which demonstrate the benefit of exploiting spatial information when aligning the domain distribution for urban scenes. The $2 \times 1$ (\ie, a top region and a bottom region) gains improvement when compared to the baseline, largely because it uses the semantic prior(\eg, the \textit{sky} is in the top region whereas the \textit{road} is in the bottom region). $2 \times 2$ gets little further improvement over $2 \times 1$,  mainly because that the urban scene is generally vertically symmetric, thus a further partitioning the regions along the vertical middle does not help too much. The $3 \times 3$ partition gives the best results, a possible explanation is that it contains a central region, which further copes with the size variance caused by the perspective transformation (\eg, the objects in the central region are smaller, while the objects outside the central are larger).

\begin{table*}
\center
\resizebox{\textwidth}{!}{%
\setlength{\tabcolsep}{3pt}
\begin{tabular}{ c | c | c c c c c c c c c c c c c c c c c c  c | c}
\hline

& & \vtext{road}& \vtext{sidewalk} & \vtext{building} & \vtext{wall} & \vtext{fence} & \vtext{pole} & \vtext{traffic light} & \vtext{traffic sign} & \vtext{vegetation} & \vtext{terrain} & \vtext{sky} & \vtext{person} & \vtext{rider} & \vtext{car} & \vtext{truck} & \vtext{bus} & \vtext{train} & \vtext{motorbike} & \vtext{bicycle} & \vtext{mean IoU} \\ \hline 

\multirow{6}{*}{\vtext{GTAV}}
& NonAdapt \cite{hoffman2016fcns}& 31.9 &  18.9&  47.7& 7.4 & 3.1 & 16.0 & 10.4 & 1.0& 76.5 & 13.0 & 58.9 & 36.0 & 1.0 &  67.1& 9.5 & 3.7 &  0.0& 0.0 & 0.0  & 21.1 \\ \cline{2-22}
& FCNs Wld \cite{hoffman2016fcns}& 70.4 & \textbf{32.4} & 62.1 & 14.9 &  5.4 &  10.9 & 14.2 &  2.7 &  79.2 & 21.3 & 64.6 & 44.1 &  4.2 &  70.4 &  8.0 & 7.3 & 0.0 & 3.5 & 0.0 & 27.1  \\  \cline{2-22}
& NonAdapt \cite{zhang2017curriculum} & 18.1&  6.8&  64.1&  7.3 & 8.7 & 21.0 & 14.9 & \textbf{16.8} & 45.9 & 2.4 & 64.4 & 41.6 & \textbf{17.5} & 55.3 & 8.4 & 5.0 & 6.9 & 4.3 &  13.8 & 22.3 \\ \cline{2-22}
& Curriculum \cite{zhang2017curriculum}& 74.8 & 22.0& 71.7  & 6.0 &  11.9  & 8.4 & 16.3   & 11.1  &75.7 & 13.3& 66.5& 38.0& 9.3  & 55.2  & 18.8 & 18.9 & 0.0 & \textbf{16.8} & \textbf{14.6} &28.9 \\ \cline{2-22}
& NonAdapt & 29.8 & 16.0 & 56.6 &  9.2 &  17.3 &  13.5 & 13.6 & 9.8 & 74.9 & 6.7 & 54.3 & 41.9 &  2.9 &  45.0 & 3.3 &  13.1 & 1.3 & 6.8 & 0.0 & 21.9 \\ \cline{2-22}
& Ours & \textbf{85.4}  & {31.2}  & \textbf{78.6} & \textbf{27.9}  & \textbf{22.2}  & \textbf{21.9 } & \textbf{23.7}  & {11.4}  & \textbf{80.7}  & \textbf{29.3} & \textbf{68.9} & \textbf{48.5} & {14.1}  & \textbf{78.0}  & \textbf{19.1}  & \textbf{23.8}  & \textbf{9.4}  & {8.3}  & 0.0 & \textbf{35.9} \\  \hline \hline

\multirow{6}{*}{\vtext{SYNTHIA}}
& NonAdapt \cite{hoffman2016fcns}& 
6.4 & 17.7 & 29.7 & 1.2 & 0.0 & 15.1 & 0.0 &  7.2 & 30.3 & - & 66.8 & 51.1 &
 1.5 & 47.3 & - &  3.9 & - & 0.1 & 0.0 & 17.4 \\ \cline{2-22}

& FCNs Wld \cite{hoffman2016fcns}& 
11.5 & 19.6 & 30.8 & 4.4 & 0.0 & 20.3 & 0.1 & 11.7 & 42.3 & - & 68.7 & 51.2 &
 3.8 & 54.0 & - &  3.2 & - & 0.2 & 0.6 & 20.2  \\  \cline{2-22}

& NonAdapt \cite{zhang2017curriculum}&  
 5.6 & 11.2 & 59.6 & 0.8 & \textbf{0.5} & 21.5 & 8.0 &  5.3 & 72.4 & - & 75.6 & 35.1 &
 9.0 & 23.6 & - &  4.5 & - & 0.5 & 18.0& 22.0 \\  \cline{2-22}

& Curriculum \cite{zhang2017curriculum}& 
65.2 & 26.1 & 74.9 & 0.1 & \textbf{0.5} & 10.7 & 3.7 &  3.0 & 76.1 & - & 70.6 & 47.1 &
 8.2 & 43.2 & - & \textbf{20.7} & - & 0.7 & 13.1& 29.0\\  \cline{2-22}

& NonAdapt & 
 4.7 & 11.6 & 62.3 & \textbf{10.7}& 0.0 & 22.8 & 4.3 & 15.3 & 68.0 & - & 70.8 & \textbf{49.7} & 
 6.4 & 60.5 & - & 11.8 & - & 2.6 & 4.3 &25.4 \\  \cline{2-22}&
Ours & 
\textbf{77.7} & \textbf{30.0} & \textbf{77.5} &  9.6& 0.3 & \textbf{25.8} &\textbf{10.3} & \textbf{15.6} & \textbf{77.6} & - & \textbf{79.8} & 44.5 & 
\textbf{16.6} & \textbf{67.8} & - & 14.5 & - &  \textbf{7.0}&\textbf{23.8} &\textbf{36.2} \\  \hline
\end{tabular}
}
\vspace{2mm}
\caption{Comparison with state-of-the-arts methods for semantic segmentation on Cityscapes using synthetic datasets as the training data. \textbf{Top:} adapting from GTAV, \textbf{Bottom:} adapting from SYNTHIA. Results of state-of-the-art methods are from their papers. We use VGG-16 as the backbone network for fair comparison. The best results are denoted in bold.}
\vspace{-0mm}
\label{tab:stat_of_the_arts}
\end{table*}

\subsection{Comparison to State-of-the-arts}
\label{sec:exp_sota}
In this section, we compare our method with two recent methods on semantic segmentation of urban scenes. The first one is \textit{FCNs in the Wild}~\cite{hoffman2016fcns}, which also adopted an adversarial training strategy, and further deployed a category specific adaptation method to align pixels of two domains that are likely from the same category. The second one is the \textit{curriculum learning adaptation}~\cite{zhang2017curriculum}, in which a curriculum learning approach was used to progressively adapt to the target domain. In both works, the dilation model with VGG-16 backbone was used as the base network. 

For a fair comparison, we compare the results of our ROAD-Net based on VGG-16 model with those two state-of-the-arts, which are summarized in the top part of Table~\ref{tab:stat_of_the_arts}. The ``NonAdapt'' baselines for each work are also included for comparison. We observe that, the results from our ``NonAdapt'' baseline is similar to those reported in \cite{hoffman2016fcns,zhang2017curriculum}, despite that we use the DeepLab model while Dilation network is used their experiment. Moreover, by using the proposed two module for real style orientation and real distribution orientation, our ROAD-Net achieves a mean IoU of $35.9\%$, which outperforms the two state-of-the-art methods. Specifically, we achieves $+7.0\%$ and $+8.8\%$ improvement compared to \cite{zhang2017curriculum} and \cite{hoffman2016fcns} respectively.

\subsection{Additional Results on SYNTHIA}
To further validate the effectiveness of the proposed method, we additionally perform an experiment using SYNTHIA~\cite{ros2016synthia} as source domain and Cityscapes as target domain. SYNTHIA is a dataset with synthetic images of urban scenes, with pixel-wise annotations. The rendering are across a variety of environments and weather conditions. In our experiment, we use the SYNTHIA-RAND-CITYSCAPES subset, which contains 9,400 images compatible with the cityscapes classes. All experiment settings remain unchanged with the previous experiments. For fair comparison, VGG-16 is used as backbone model. The results of all methods are summarized in the bottom part of Table~\ref{tab:stat_of_the_arts}. The ``NonAdapt" baseline for each method is also included for comparison. We observe that our proposed approach outperforms the other methods by a large margin, which again demonstrates the effectiveness of our ROAD Net in cross-domain semantic segmentation of urban scenes.

\section{Conclusion}
In this paper, we have presented a new model \textit{Reality Oriented ADaptation Network}(ROAD-Net) for semantic segmentation of urban scenes by learning from synthetic data. Two modules are proposed in our paper, \textit{target guided distillation} and \textit{spatial-aware adaptation}, where the former one aims to adapt the style from the real images by imitating the pretrained network, and the latter one is used to reduce the domain distribution mismatch with the help of layout information. Those two modules can be integrated with different semantic segmentation networks to improve their generalizability when applying to a new domain. We evaluate the proposed method on Cityscapes, using GTAV and STYNHIA as the source domain. The experiments on benchmark datasets have clearly demonstrated the effectiveness of our proposed ROAD-Net.

\paragraph{Acknowledgements}
Wen Li is the corresponding author. This project is supported in part by Toyota Motor Europe. The authors gratefully thank NVidia Corporation for donating GPUs.

{\small
\bibliographystyle{ieee}
\bibliography{cvpr2018}
}

\end{document}